\documentclass[conference]{IEEEtran}
\IEEEoverridecommandlockouts
\pdfoutput=1
\usepackage{cite}
\usepackage{amsmath,amssymb,amsfonts}
\usepackage{algorithmic}
\usepackage{dblfloatfix}
\usepackage{graphicx}
\usepackage{textcomp}
\def\BibTeX{{\rm B\kern-.05em{\sc i\kern-.025em b}\kern-.08em
    T\kern-.1667em\lower.7ex\hbox{E}\kern-.125emX}}
\begin{document}

\title{A Convolutional Neural Network based Live Object Recognition System as Blind Aid}

\author{\IEEEauthorblockN{Kedar Potdar}
 \IEEEauthorblockA{Student, New York University \\
  New York City,\\
  NY USA 11209 \\
  kedar1916@gmail.com}
 \and
 \IEEEauthorblockN{Chinmay D. Pai}
 \IEEEauthorblockA{Student, Computer Engineering \\
  WIEECT, Worli\\
  Mumbai, INDIA, 400 018 \\
  chinmaydpai@gmail.com}
 \and
 \IEEEauthorblockN{Sukrut Akolkar}
 \IEEEauthorblockA{Student, Computer Engineering \\
  WIEECT, Worli\\
  Mumbai, INDIA, 400 018 \\
  sukrutakolkar@gmail.com}
}

\maketitle

\begin{abstract}
 This paper introduces a live object recognition system that serves as a blind aid. Visually impaired people heavily rely on their other senses such as touch and auditory signals for understanding the environment around them. The act of knowing what object is in front of the blind person without touching it (by hands or using some other tool) is very difficult. In some cases, the physical contact between the person and object can be dangerous, and even lethal.

 This project employs a Convolutional Neural Network for recognition of pre-trained objects on the ImageNet dataset. A camera, aligned with the system’s pre-determined orientation, serves as input to a computer system, which has the object recognition Neural Network deployed to carry out real-time object detection. Output from the network can then be parsed to present to the visually impaired person either in form of audio or Braille text.
\end{abstract}


\section{Introduction}
The white cane is the most widely used blind aid tool serving mobility purposes to the visually impaired. With advancements in technology, new tools are being developed which assist the visually impaired not only with mobility, but also with better understanding of the surroundings. This paper introduces one such system that utilizes Object Recognition with Machine Learning to help blind persons recognize objects in vicinity.

The act of recognizing objects without vision is a difficult task. Blind people primarily rely on their other senses for this purpose. The act of knowing what object is in front of the blind person without touching it (by hands or using some other tool) is very difficult. In some cases, the physical contact between the person and object can be dangerous, and even lethal. A computerized system that recognizes objects and gives either auditory or Braille feedback can help to great extents by giving a more accurate understanding of the environment and eliminating threats involved in the process.

Convolutional Neural Networks (CNNs) have shown promising results in the field of image classification\cite{b1}. The proposed system employs a Deep CNN to recognize objects using input from a camera and processing on a portable computer system. In this paper, we describe the system’s architecture and functionality.

\section{Literature Review}
Hsieh Chi-Sheng, in Electronic Walking System for the Blind\cite{b2}, have developed an electronic talking stick which alerts the visually impaired about the obstacles in front of them. The system uses optical sensors to detect the obstacles, the detected signals are then converted into voice using a control circuit. The voice output is provided through earphones attached to the handle of the stick. The scanning device is designed to detect obstacles within a range of 1 metre.

In Portable Blind Aid Device\cite{b3}, Eugene Evanitsky, Xerox Corporation, put forth a portable blind aid device based on a cell phone. The images captured by the phone are analysed to classify the spatial relationships related to moving objects. The information about the environment around the person is provided in the form of audio output.

In Intelligent Glasses For The Visually Impaired\cite{b4}, Humberto Orozeo Cervantes invented a system which describes the environment around a blind person using image analysis. Images are captured using a camera and sensor mounted on glasses, the images are sent to a remote server machine for analysis. The audio output is sent back to the user’s output device.

A. Jothimani, Shirly Edward, and G. K. Divyashree have described a device for that performs object detection for the visually impaired in Object Identification for Visually Impaired\cite{b5}. The system employs coarse description technique to detect objects in images captured by a camera module.

Jamal. S. Zraquo, Wissam M. Alkhadour and Mohammad Z. Siam, propose an object recognition system based on SURF feature descriptor in Real-Time Objects Recognition Approach for Assisting Blind People\cite{b6}. The system uses a two-camera setup and an ultrasonic sensor along with GPS for depth-mapping and localization. Features extracted from the images using SURF feature descriptor\cite{b7} are verified using a model database.

\section{Object Detection}
Object detection is one of the classical problems in the field of computer vision. Object Detection technically deals with recognition of objects of greater importance in the input data and detecting their location in the object space. Detection of an object in a spatial relation is a far more intricate problem than classifying the detected objects, since it also provides location of the object in the input data, wherein the classifier fails to foster any competence. A typical object detection algorithm aims to extract the features from the input data and apply learning methods to recognize and illustrate the instances of objects categorically. During extraction, the input is scanned for sets of objects, or features which are constituted in the data. These features are then further scanned for regional bounding, and the boundaries for the features are computed using boundary detection algorithms and the boundary output is then used as the location for the objects in the space. This location along with the detected object is passed through a neural network to compute the features, which is then classified by the network.

Boundary detection algorithms are specifically used to detect the objects by moving a smaller frame over the object, sliding through the entire image and finding orientations. Each orientation computed in a window is then analyzed for any significant differences between multiple frames, and the boundary region is computed. The Neural Network tries to compute the features in the extracted object regions and classifies them categorically. This classification is scored on a predefined scale in the network based on confidence of the classified object.

\section{Convolutional Neural Network for Object Detection}

\subsection{Convolutional Neural Networks}
Convolutional Neural Networks are Artificial Neural Network models that are comparable to visual cortex, which is a fully connected, stratified network with each layer providing regions of cells that are sentient of specific fields of vision. This type of networks are specifically designed for making full use of data with good spatial relation. The stratified topology of the convolutional neural network wheels out different properties of each layers, serialized as a collection of convolutional, activation, pooling and fully connected layers, with the entire network comprising of at least one layer of convolution. This convolutional layer takes advantage of the fact that input is made up of spatially related data, and have neurons arranged in 3 dimensions consisting of width, height and depth of the activation volume. Mathematically, this layer computes a dot product between the weights of local receptive fields in the input and the connected region in the input volume, and produces another array of numbers known as activation map or feature map. Each local receptive field converges into a hidden neuron connected to it in the succeeding layer.

\begin{figure*}[htbp]
 \centerline{\includegraphics[width=\textwidth, height=4cm]{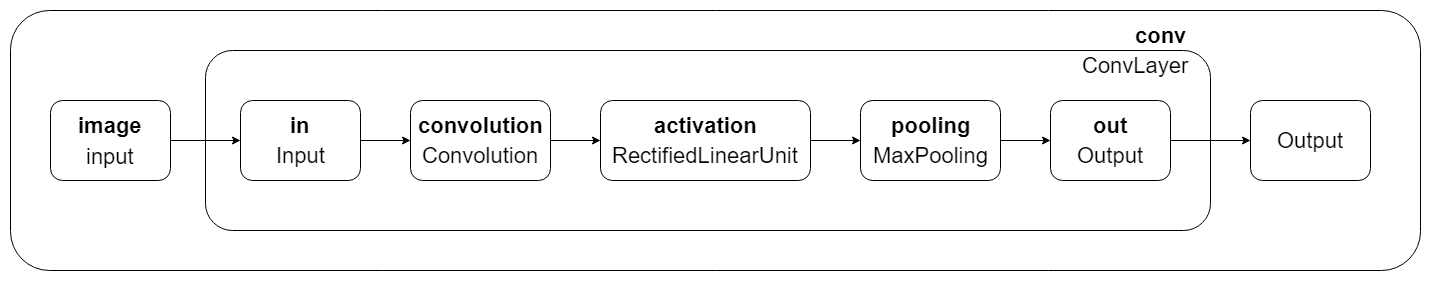}}
 \caption{CNN baseline architecture.}
 \label{cnn_arch}
\end{figure*}

Since the output of the convolution produces a linear transformation of the input, it does not satisfy the universal approximation theorem which insinuates that the representational power of the network is coerced with linearity. Hence, for the network to comply with the universal approximator, activation layer is required. The sole purpose of the activation layer is to infuse non-linearities in the network.

The output of this layer is then pooled or downsampled in the pooling layer to simplify the feature map produced by the preceding layers. The pooling layer takes advantage of the fact that when a feature is extracted by the previous layers, the location of the feature in the feature map is not as important as its location comparable to other detected features. This reduces the spatial size of the representation, consequently reducing the computational cost of the entire network and abating overfitting as well.

The entire composed stack is then consolidated into a fully connected layer, wherein each neuron is connected to every neuron in the preceding layer. The fully connected layer is a usually one-dimensional and comprises all the labels that are to be classified. This layer outputs a score for each label of classification.

Since the convolutional networks are specifically designed to make use of spatial relations between the objects in the provided data, they perform better in regions of machine learning where the related spatial data needs to be manipulated, such as detection or recognition of objects. Also, as these convolutional networks explicitly assume that the inputs are images, it provides a streamlined function to encode the data and implement the network with immensely reduced parameters.

\subsection{Image Recognition}
Classification of objects in an image is a simple task as long as it pertains to the humankind, but it is quite an involved and elusive process for machines. One of the non-exhaustive crucial challenges that a machine fails to deal with is the semantic aspect relating to the recognition and classification of visual data, which can be easily determined by humans. Image classification helps the machine identify the features of the image and categorize them into their appropriate class. The process of image classification is diverse and comprises of image acquisition, pre-processing, detection, extraction and classification. During pre-processing, the images are transformed, denoised and analyzed for location of main component data. This image is then passed through detection and feature extraction so as to deduce the objects of significance. The features classified based on the extracted features are categorized in accordance to the predefined classes in the image database.

\subsection{ImageNet Inception Model}
The ImageNet dataset consists around 15 million labelled images classified into approximately 22000 categories\cite{b8}. This wide set of data can be put to a great use to describe spatial relationships between objects and their location in the environment. Our network is tightly based around the ImageNet dataset and Deep Convolutional Neural Networks for image classification\cite{b9}.

Image classification using Neural Networks is a very strenuous task. Convolutional Neural Networks are models which are specifically designed to provide layers, where each layer is capable of processing information pertaining field of vision provided as an input. Deep convolutional neural networks using supervised learning has been proven to be revolutionary for image classification\cite{b9, b10, b11}. By stacking layers on top of each other, a deep network is pivotal in providing better results as

compared to other networks, as more layers induce increased quality of features in the data. This paper addresses image classification in real-time in a Three-dimensional space using Deep Convolutional Neural Network\cite{b9} and ImageNet dataset\cite{b8} for an object recognition system that serves as an aid for the visually impaired.

\section{System Architecture}
The system comprises of a portable computational hardware module and a neural network model using Convolutional Neural Networks. This section describes the system's hardware and software architecture.
\subsection{Hardware Overview}
\begin{figure}[htbp]
 \centerline{\includegraphics[width=8cm, height=8cm]{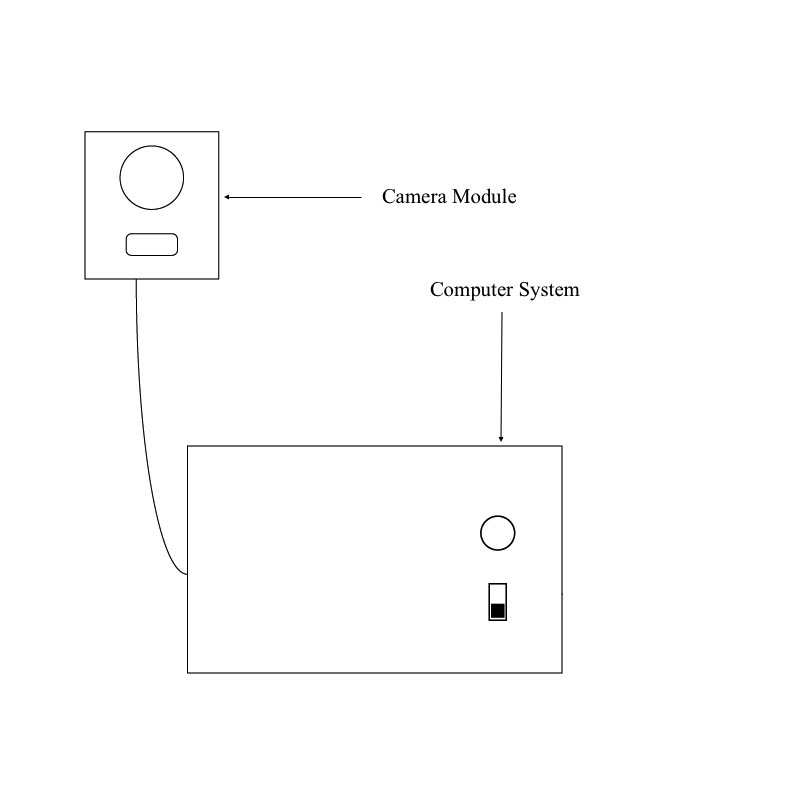}}
 \caption{System architecture.}
 \label{sys_arch}
\end{figure}

The computational hardware consists of a camera and a computer system which employs the neural network. Digital images are transferred from the camera to the computer system for processing. ``Fig. ~\ref{sys_arch}''. shows the assembly of the system.

\subsection{Software Overview and Network Architecture}
The system utilizes Python programming language with Google's Tensorflow Machine Learning Library\cite{b12} to build and deploy the CNN. The performance is analyzed based on real world scenarios tested on the neural network. The network architecture makes use of 9 convolutional and max-pooling layers, followed by 2 fully connected layers. The network baseline architecture is shown in ``Fig.~\ref{cnn_arch}''. The network is a concoction of classification and detection models.

\section{Methodology}
The system images on the press of a button and the captured data is transmitted to the computational module wherein the data is analyzed, preprocessed and computed for object detection and classification. The entire process is illustrated in the flowchart shown in figure. This section describes each phase in the process.

\subsection{Preprocessing}
Detection of objects requires well-detailed data input, and at the same time be less computationally intensive. Hence, we resize the input data resolution to 416 x 416. This ensures that the information is preserved and the computational overhead for the portable hardware is reduced.

\begin{figure}[!b]
 \centerline{\includegraphics[width=1.75cm, height=6.5cm]{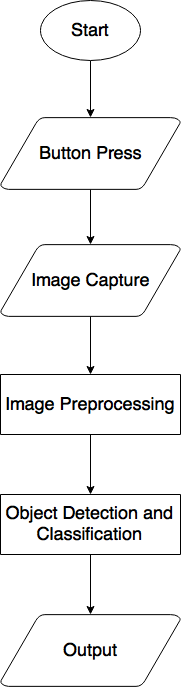}}
 \caption{System flowchart.}
 \label{flowchart}
\end{figure}

\begin{figure*}[htbp]
 \centerline{\includegraphics[width=\textwidth, height=4cm]{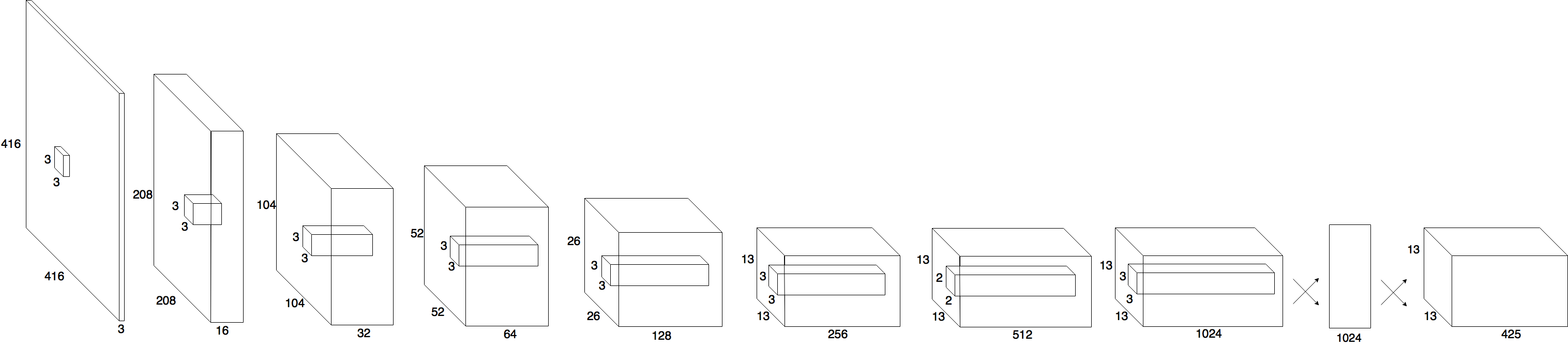}}
 \caption{Complete network architecture.}
 \label{net_arch}
\end{figure*}

\subsection{Classification and Detection}
The classification network uses TensorFlow Models trained on the ImageNet 1000-class competition dataset. TensorFlow is an interface for expressing Machine Learning algorithms, and an implementation for executing such algorithms \cite{b12}. This network model is inspired by the YOLO Image Classification model \cite{b13} and to train the model we use 7 convolutional layers comprising of a max-pooling layer between each layer, followed by a fully connected layer. The trained data is cross-validated using the ImageNet validation dataset. The network is then modified to accommodate object detection layer, which as stated by Ren et al. that adding both convolutional and connected layers to pertained networks can increase the network performance \cite{b14}. The final fully connected layer forecasts categorical probability as co-ordinates of the detected object. The co-ordinates for the detected objects are normalized between 0 to 1 so as to reduce the computational complexity of the network.

The entire network uses a variation of ReLU activation function, exponential linear units (ELU) to make the mean activations closer to zero to increase the learning speed \cite{b15}. Exponential Linear Units can be defined as:

\begin{equation}
 {\displaystyle f(x)={\begin{cases}x&{\mbox{if }}x>=0\\a[\exp(x)-1]&{\mbox{otherwise}}\end{cases}}}
\end{equation}

Where \(a\) is the hyper-parameter to be tuned and \(a>=0\) is a constraint.

\begin{figure}[htbp]
 \centerline{\includegraphics[width=8cm, height=5cm]{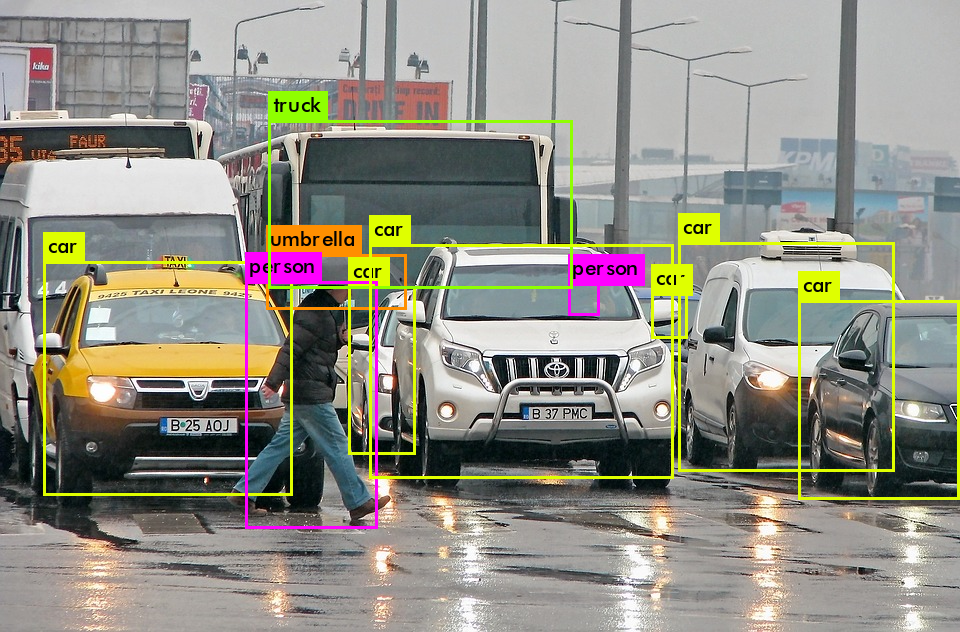}}
 \caption{Object detection in a dense scene.}
 \label{road_res}
\end{figure}

Our implementation requires the network to have high computation speed with good accuracy. Hence, ELU is preferably used over ReLU as it provides higher classification speed and accuracy \cite{b15}. The entire network architecture is shown in ``Fig.~\ref{net_arch}''.

\begin{figure}[!b]
 \centerline{\includegraphics[width=8cm, height=5cm]{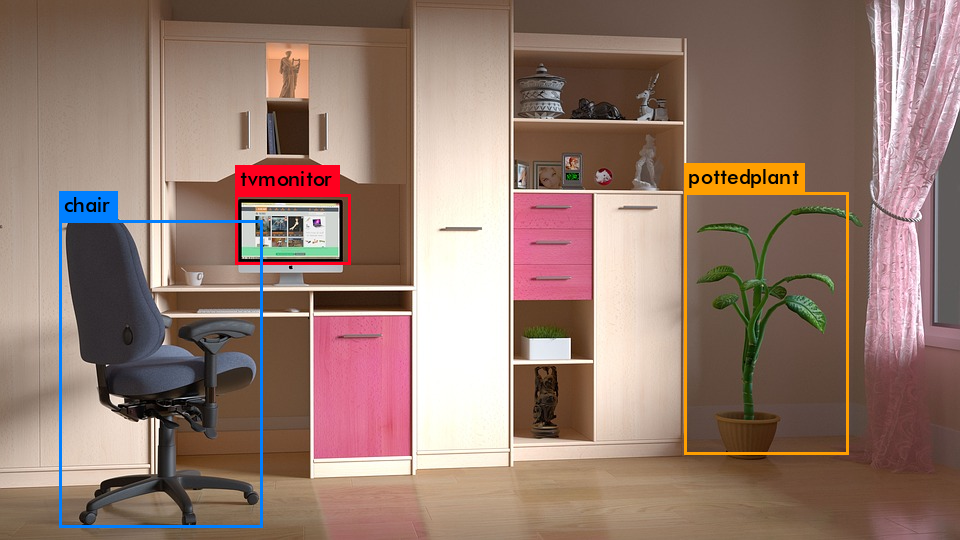}}
 \caption{Object detection in an indoor environment.}
 \label{home_res}
\end{figure}

\section{Results}

We tested our model on the ImageNet 200-class validation dataset from ILSRVC 2014 challenge. The model achieves 50 mAP in 200 object categories and a top-1 accuracy of 70.6\% from the ImageNet validation dataset. Since the network model requires a single evaluation, the computation time is quite low, which is desirable to perform real time object detection.

Two resultant images are illustrated in ``Fig.~\ref{road_res}'' and ``Fig.~\ref{home_res}''. In ``Fig.~\ref{road_res}'', the model is able to accurately detect the cars and a person crossing the road. Most of the objects in the scene are recognized correctly, even the ones that are not completely visible. This shows that the model performs well in cluttered scenes. Meanwhile, ``Fig.~\ref{home_res}'' is an image captured indoors, where the model is able to correctly identify the objects of interest. Albeit, the accuracy suffers when it comes to recognizing smaller objects or those in lower resolutions. For example, the bus is recognized as a truck in ``Fig.~\ref{road_res}''. The accuracy of the model can be improved by increasing the number of convolutional and pooling layers. But since the model is running on a portable device, increasing the levels of computation directly impacts the computation time, which at the time of writting this paper is a current limitation of the system.

\section{Conclusion and Future Work}

We discussed a live object recognition system using convolutional neural networks that serves as a blind aid. Our implementation of the model demonstrates that such a system can be used to identify objects and help the visually impaired. The results show that the recognition achieved by the system is fairly accurate, albeit it fails to identify smaller, trivial objects. The computation time required for detection is quite low while having a decent accuracy on a portable device, which shows that the system can be used to interpret objects in the environment in real time. The model produces the output in a JSON format, which can be relayed in auditory format or braille texts. Further improvements can be done to improve the accuracy of the system by adding more convolutional layers at the cost of increase in computation time, which is a prevailing barrier for portable systems. Also, improvements can be made by enhancing the images before detecting the features. In future, we plan on improving the accuracy and adding depth recognition to the system to help locate the object in a spatial domain in real time.

\end{document}